\documentclass[10pt,twocolumn,letterpaper]{article}

\usepackage{iccv}  

\usepackage{graphicx}
\usepackage{caption}
\usepackage{tipa}
\usepackage{blindtext}
\usepackage{lineno}
\usepackage{colortbl}
\usepackage{amsmath,amssymb,amsfonts}
\usepackage{algorithmic}
\usepackage{graphicx}
\usepackage{multirow}
\usepackage{adjustbox}
\usepackage{amsmath}
\usepackage{wasysym}
\usepackage{footnote}
\usepackage{booktabs}
\usepackage{amssymb}
\usepackage{utfsym}
\usepackage{float}

\definecolor{iccvblue}{rgb}{0.21,0.49,0.74}
\usepackage[pagebackref,breaklinks,colorlinks,allcolors=iccvblue]{hyperref}

\definecolor{teaserred}{RGB}{180,10,56}

\definecolor{teaserblue}{RGB}{0,15,139}

\definecolor{uclablue}{RGB}{159, 195, 224}

\definecolor{uclagold}{RGB}{254,180,167}

\definecolor{grayred}{RGB}{232,237,205}

\usepackage{color, colortbl}

\definecolor{iccvblue}{rgb}{0.21,0.49,0.74}
\definecolor{TealBlue}{rgb}{1.0, 0.97, 0.8}

\def\paperID{****} 
\def\confName{ICCV}
\def\confYear{2025}

\title{WikiAutoGen: Towards Multi-Modal Wikipedia-Style \\ Article Generation}

\author{
    Zhongyu Yang\textsuperscript{\rm 1, 2}\footnotemark[1]\, \footnotemark[2],
   Jun Chen\textsuperscript{\rm 3, 1}\footnotemark[1], Dannong Xu\textsuperscript{\rm 1, 4} \footnotemark[2], Junjie Fei\textsuperscript{\rm 1} \\
  Xiaoqian Shen\textsuperscript{\rm 1},
  Liangbing Zhao\textsuperscript{\rm 1},
  Chun-Mei Feng\textsuperscript{\rm 5}, Mohamed Elhoseiny\textsuperscript{\rm 1} \\
  \textsuperscript{\rm 1}{King Abdullah University of Science and Technology},\,    \\
  \textsuperscript{\rm 2}{Lanzhou University} \,    
  \textsuperscript{\rm 3}{Meta AI} \,   
  \textsuperscript{\rm 4}{The University of Sydney} \,   
  \textsuperscript{\rm 5}{IHPC, A*STAR} \\
    \parbox[c]{\textwidth}{\centering  
    \texttt{\{zhongyu.yang, jun.chen, junjie.fei, xiaoqian.shen\\
     liangbing.zhao, mohamed.elhoseiny\}@kaust.edu.sa}}\\
  \texttt{daxu8019@uni.sydney.edu.au}, \texttt{fengcm.ai@gmail.com}
}

\begin{document}

\twocolumn[{
\renewcommand\twocolumn[1][]{#1}
\vspace{-20pt}
\maketitle
\vspace{-25pt}
\begin{center}
    \captionsetup{type=figure}
    \centering
    \includegraphics[width=\textwidth]{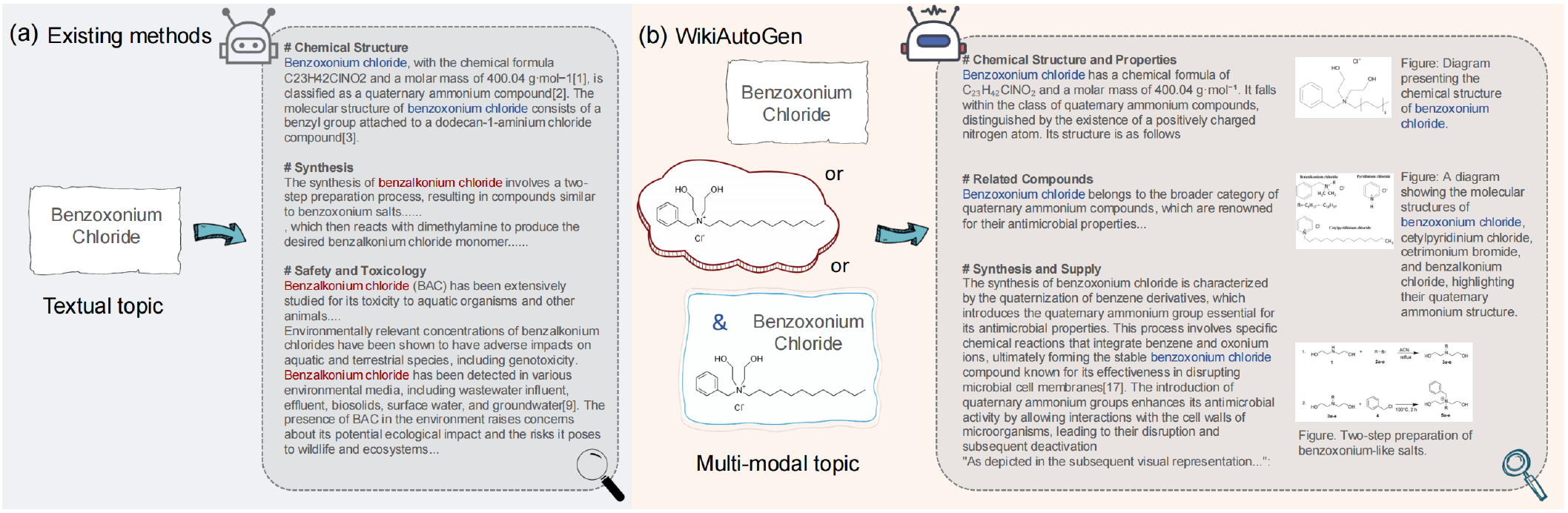}
    \caption{\textbf{Comparison of existing text-only article generation methods and our proposed WikiAutoGen.}  
    Existing approaches~\cite{Shao2024AssistingIW, Jiang2024IntoTU} rely exclusively on textual sources, often producing inconsistent or inaccurate results. For example, in \textbf{(a)}, the target topic is \textcolor{teaserblue}{\textit{‘Benzoxonium Chloride’}}, yet the baseline incorrectly generates information about \textcolor{teaserred}{\textit{‘Benzalkonium Chloride’}}. In contrast, our \textbf{WikiAutoGen} framework integrates both visual and textual modalities to generate coherent multimodal content. Additionally, WikiAutoGen employs a multi-perspective self-reflection mechanism, significantly improving content accuracy and reliability, as illustrated in \textbf{(b)}.}
        \label{teaser}
\end{center}
}]

\footnotetext[1]{$^*$ Equal contribution}
\footnotetext[2]{$\dagger$ Work done during an internship at KAUST}

\maketitle
\begin{abstract}
Knowledge discovery and collection are intelligence-intensive tasks that traditionally require significant human effort to ensure high-quality outputs. Recent research has explored multi-agent frameworks for automating Wikipedia-style article generation by retrieving and synthesizing information from the internet. However, these methods primarily focus on text-only generation, overlooking the importance of multimodal content in enhancing informativeness and engagement. In this work, we introduce WikiAutoGen, a novel system for automated multimodal Wikipedia-style article generation. Unlike prior approaches, WikiAutoGen retrieves and integrates relevant images alongside text, enriching both the depth and visual appeal of generated content. To further improve factual accuracy and comprehensiveness, we propose a multi-perspective self-reflection mechanism, which critically assesses retrieved content from diverse viewpoints to enhance reliability, breadth, and coherence, etc. Additionally, we introduce WikiSeek, a benchmark comprising Wikipedia articles with topics paired with both textual and image-based representations, designed to evaluate multimodal knowledge generation on more challenging topics. Experimental results show that WikiAutoGen outperforms previous methods by 8\%-29\% on our WikiSeek benchmark, producing more accurate, coherent, and visually enriched Wikipedia-style articles.
Our code and examples are available at \url{https://wikiautogen.github.io/}
\end{abstract}    
\section{Introduction}
\label{sec:intro}

Knowledge discovery and content generation are essential for organizing and disseminating information, but they remain time-consuming and intelligence-intensive, requiring substantial human effort to collect, structure, and verify information. With the advent of large-scale AI models like large language models (LLMs)~\cite{chatgpt, gemini, llama3, qwen, deepseek}, there is growing potential to automate knowledge collection, synthesis, and organization in a more efficient and scalable manner~\cite{Jiang2024IntoTU, Shao2024AssistingIW}. Such automation not only accelerates knowledge discovery but also enhances accessibility, making information more readily available and up to date.

Recently, several methods, such as Storm~\cite{Shao2024AssistingIW} and Co-Storm~\cite{Jiang2024IntoTU}, have been proposed to automate Wikipedia-style article generation. While they can produce Wikipedia-like content, they still face key limitations: (1) they are limited to text-only generation and lack the ability to incorporate multimodal content such as relevant images;
(2) the generated articles often lack breadth, depth, and reliability, reducing their overall informativeness and credibility.

To address these challenges, we introduce WikiAutoGen, a multi-agent framework designed to generate high-quality, multimodal Wikipedia-style articles automatically. Unlike prior works, WikiAutoGen can directly search both textual and visual information and generate multimodal content (see Figure \ref{teaser}), enriching article content with relevant and diverse modalities. Additionally, we propose a novel multi-perspective self-reflection module, which enables the system to self-regulate, refine, and critically evaluate its generated content. 
This mechanism enhances the reliability, depth, and breadth of the information by encouraging iterative improvement and multi-source validation.

To advance the development and evaluation of multimodal knowledge generation, we introduce \textbf{WikiSeek}, a new benchmark designed to tackle challenging topics comprising both textual and visual components. 
Existing benchmarks primarily focus on text generation or cover only straightforward topics (see Table \ref{tab:compare}). 
In contrast, WikiSeek is multimodal and specifically targets more complex subjects with limited coverage on Wikipedia, making it significantly harder for current methods to retrieve and synthesize comprehensive information. This increases the challenge of content generation, pushing models to explore deeper, enhance their retrieval capabilities, and improve their ability to handle underexplored subjects.

\begin{table}[t!]
  \centering
\resizebox{\linewidth}{!}{
  \begin{tabular}{l|ccc}
    \toprule
    \multirow{2}[4]{*}{DATASETS} &  
    \multicolumn{3}{c}{Dataset Statistics}       \\
\cmidrule{2-4}          & Type  & Retrieval Modality & Difficulty Levels  \\
    \midrule
        Surfer100 \citep{Li2021Surfer100GS} & \includegraphics[height=1em]{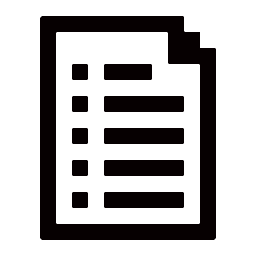}& \includegraphics[height=1em]{Fig/text.png}     &  easy    \\
         FreshWiki \citep{Shao2024AssistingIW} & \includegraphics[height=1em]{Fig/text.png}  & \includegraphics[height=1em]{Fig/text.png}   & easy \\
        IRP \citep{Balepur2023ExpositoryTG} & \includegraphics[height=1em]{Fig/text.png}  & \includegraphics[height=1em]{Fig/text.png}   & n/a     \\
        WildSeek  \citep{Jiang2024IntoTU} & \includegraphics[height=1em]{Fig/text.png}   & \includegraphics[height=1em]{Fig/text.png}  & n/a     \\
        WikiSeek (Ours) & \includegraphics[height=1em]{Fig/text.png} \includegraphics[height=1em]{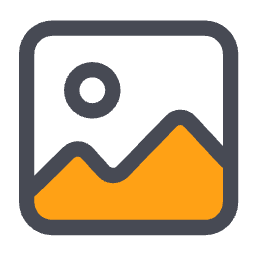}     & \includegraphics[height=1em]{Fig/text.png} \includegraphics[height=1em]{Fig/pic.png}  & high   \\
    \bottomrule
    \end{tabular}%
}
    \vspace{-3mm}
    \caption{
    \textbf{Comparison of WikiSeek with existing benchmarks.} Modalities are indicated by \includegraphics[height=1em]{Fig/text.png} (text) and \includegraphics[height=1em]{Fig/pic.png} (images). Difficulty levels are categorized based on the average number of characters in corresponding Wikipedia pages: difficult ($<$500 characters), medium (500–2000 characters), and easy ($>$2000 characters). }
  \label{tab:compare}
\vspace{-6mm}
\end{table}%

Extensive experiments demonstrate that WikiAutoGen significantly outperforms existing methods in generating high-quality textual and visual content. We evaluated text quality across nine key dimensions and image quality across four essential criteria. Experimental results show that 
WikiAutoGen outperforms prior methods by 8\%–29\% in textual quality and 11\%–14\% in image quality, demonstrating consistent gains across the input topics of text-only, image-only, and image-text tasks.

Our contributions can be summarized as follows:

\begin{itemize}
    \item We introduce WikiSeek, a new multimodal benchmark designed for evaluating Wikipedia-style article generation, featuring challenging topics with limited existing coverage, represented through both text and images.

    \item We propose WikiAutoGen, a multimodal article generation framework that synthesizes comprehensive content by effectively integrating textual and visual inputs. On WikiSeek, WikiAutoGen achieves 8\%–29\% improvements over the previous best model in textual generation.
    
    \item We develop a novel multi-perspective self-reflection module, which iteratively enhances readability, informativeness, and coherence by incorporating feedback from diverse roles, including reader, writer, and editor.
\end{itemize}

\section{Related Work}
\label{sec:relatedwork}

\noindent\textbf{Automatic Expository Writing.}~
LLMs have shown strong performance in automatic expository writing, particularly in generating Wikipedia-style articles~\cite{tan2024proxyqa,kumar2024longlamp,wu2024spinning,min-etal-2023-factscore, Besta2024DemystifyingCT, Wang2024AutoSurveyLL}.
Despite their strength in traditional Natural Language Generation (NLG), LLMs still struggle to produce long-form text that is coherent and logically structured~\cite{que2024hellobench,dong-etal-2024-bamboo,tang-etal-2023-enhancing, Shen2023BeyondSD, Wang2024AutoPatentAM}.
To address this, \cite{tan-etal-2021-progressive} proposed using domain-specific keywords that are progressively refined into full passages through multi-stage generation. Notably, Shao et al.~\cite{Shao2024AssistingIW, Jiang2024IntoTU} highlighted the crucial role of pre-writing strategies, identifying them as a key factor in improving article quality.
Recently, some automatic writing systems expanded their knowledge boundaries through mindmaps and tree-based methods~\cite{Balepur2023ExpositoryTG, Liang2024IntegratingPI, Shen2023BeyondSD, Wen2023MindMapKG}.
However, while these iterative content planning methods effectively leverage widely accessible information for common topics, they remain largely ineffective in less-explored or niche domains due to the scarcity of structured prior knowledge~\cite{Gao2023EvaluatingLL, Xi2025OmniThinkEK}.
Meanwhile, existing methods are limited to generating text-based articles and fail to incorporate other modalities, such as visual elements.
This inherent constraint in data input inevitably leads to incomplete information and reduced readability, making knowledge acquisition more challenging.
Conversely, our WikiAutoGen is the first multimodal writing system that integrates visual content and retrieves knowledge across multiple modalities, allowing visual information to complement textual content by capturing details that may otherwise be overlooked.

\noindent\textbf{Self-reflection.}~Recent progress in optimizing LLMs through self-reflection mechanisms is significant. 
The core idea is to enable models to analyze and refine their outputs through self-generated feedback.
Existing approaches construct feedback sources from three strategies:
(1) The LLM conducts iterative self-evaluation~\cite{Shinn2023ReflexionLA};
(2) A separately trained critic module provides specialized feedback~\cite{Gou2023CRITICLL};
(3) External knowledge from sources like Wikipedia and browsers is integrated~\cite{Madaan2023SelfRefineIR, Asai2023SelfRAGLT}.
Specifically, REFINER~\cite{Paul2023REFINERRF} demonstrated that a trained critic module can enhance reasoning without fine-tuning the reasoning module, supporting feedback mechanism optimization.
Further, some methods~\cite{coyne-2023-template, Zhou2024HypothesisGW} introduced feedback mechanisms based on error-type templates and context-hypothesis mappings.
Recent studies~\cite{Pich2024LLMsCL, Wang2024TasTeTL, Sun2023PrincipleDrivenSO, Wang2024ATU, Zhang2024InContextPL} focus on LLMs' in-context learning.
They design prompt templates to help models generate feedback from historical outputs or patterns.
However, the multi-perspective self-reflection method proposed by~\cite{Yan2024MirrorAM}, which relies on a Navigator-Reasoner heuristic interaction, is limited to closed-loop LLM discussions without external knowledge acquisition, resulting in restricted information richness and verifiability.
In contrast, our approach enhances multi-perspective self-reflection with multi-web search, addressing complex topic exploration and retrieval challenges while enabling multi-dimensional control over topic and article quality.


\section{WikiSeek benchmark}
\label{benchmark}

\subsection{Task Definition}
\label{sec:task_definition}
Given a topic as text ($T$), image ($I$), or a combination of both ($T, I$), the objective is to generate a Wikipedia-style article ($A$) that integrates relevant knowledge and is supported by verifiable references ($R$). This task is particularly important in domains such as investigative journalism, scientific research, and market analysis, where the generation of accurate, well-sourced content is essential.

\subsection{WikiSeek Construction}
A key challenge in this task is the lack of a suitable benchmark that effectively encompasses multimodal topics. 
Existing benchmarks, however, remain largely text-centric, thus failing to adequately reflect the complexity of multimodal content generation. 
To address this limitation, we introduce WikiSeek, a new benchmark specifically designed to evaluate more challenging topics that incorporate both text and images. WikiSeek establishes a robust evaluation framework, enabling a more comprehensive and reliable assessment of multimodal knowledge generation in practice. In the following sections, we detail the construction process of this benchmark.

\begin{figure*}[ht]
    \centering
    \vspace{-6mm}
    \includegraphics[width=\linewidth]{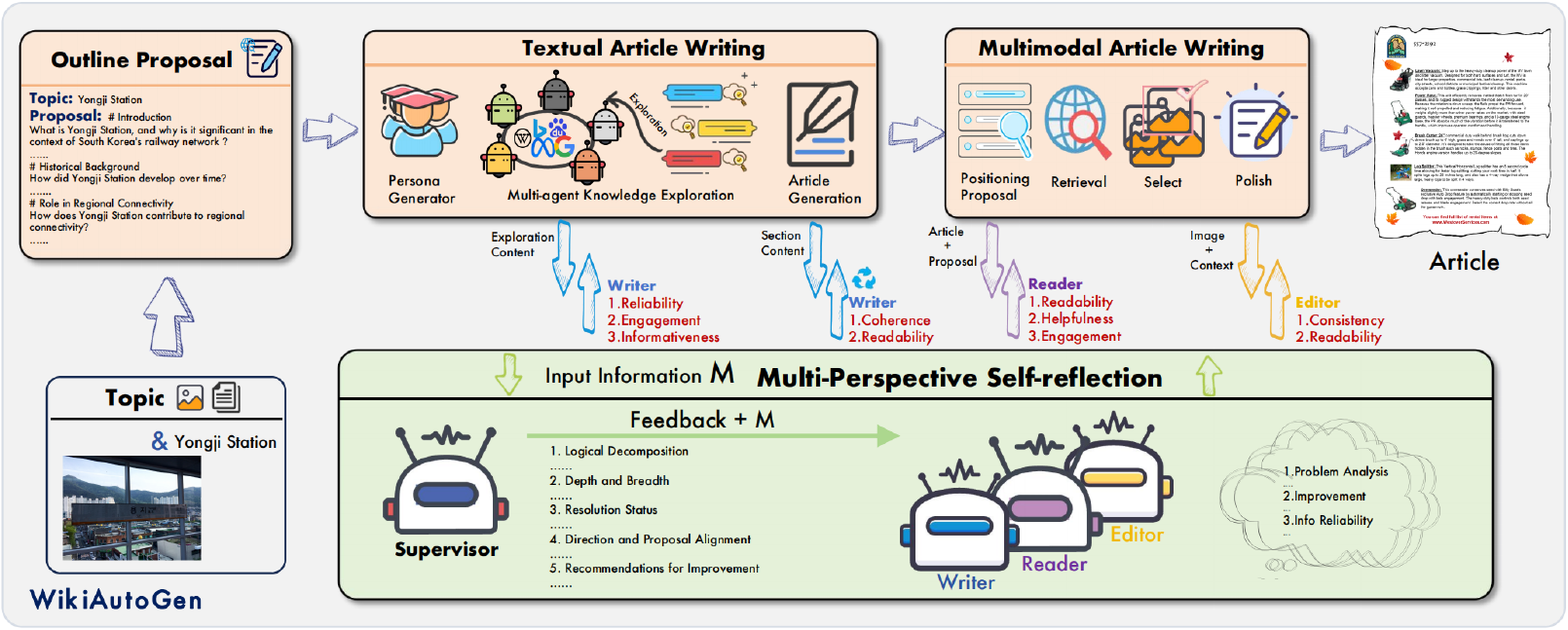}
        \caption{\textbf{Overview of WikiAutoGen}, our multimodal framework for Wikipedia-style article generation. The pipeline includes three main stages: (1) an \textbf{Outline Proposal} module that structures the article outline based on the multimodal topic input (image and text); (2) a \textbf{Textual Article Writing} module involving persona generation, multi-agent collaborative exploration, and article drafting; and (3) a \textbf{Multimodal Article Writing} module that incorporates relevant images through positioning proposals, retrieval, selection, and final polishing. The entire generation process is enhanced by a \textbf{Multi-Perspective Self-Reflection} module, leveraging supervisory and agent-specific feedback (writer, reader, editor) to iteratively improve article quality in terms of coherence, readability, and engagement, etc.}
    \label{fig:pipeline}
    \vspace{-3mm}
\end{figure*}

\noindent\textbf{Benchmark construction pipeline.} Our WikiSeek benchmark is designed with two key objectives:  
(1) to evaluate multimodal article generation where both the input and output include text and images; and (2) to target underexplored topics on Wikipedia that present greater challenges for retrieval and synthesis.

We select topics from the WikiWeb2M dataset ~\cite{Burns2023WikiWeb2MAP}, which comprises approximately 2 million English Wikipedia articles containing both text and images. To identify challenging topics, we focus on articles where the main content has fewer than 500, 300, and 100 characters, categorizing them as hard, very hard, and extremely hard, respectively. Typically, these more challenging or rare topics have minimal coverage on Wikipedia, representing less-documented and lesser-known subjects.

\noindent \textbf{Topic filtering and quality control.} To curate a high-quality benchmark that includes both challenging and meaningful topics, we implement a rigorous topic filtering process. First, we retain only Wikipedia articles with fewer than 500 characters, ensuring a focus on underexplored and more difficult topics. We then sample hundreds of topics associated with images from the WikiWeb2M dataset~\cite{Burns2023WikiWeb2MAP}.  

However, some topics, such as ``1997 in Japan" and ``.bh", are either overly general or semantically underspecified, making them unsuitable for evaluation. 
To address this, we manually verify all selected topics and remove those that lack meaningful content or pose evaluation challenges.  
After this filtering process, we obtain a final set of 300 topics, evenly distributed across three difficulty levels (hard, very hard, and extremely hard), with 100 topics per level. These topics are represented in one of three formats: text-only, image-only, or a combination of both.

\section{Method}
\label{method}

Writing high-quality multimodal Wikipedia-style articles usually requires a coordinated multi-agent system that effectively breaks down the process into distinct stages, including outline generation, web-based material retrieval, and article synthesis.
Beyond these stages, maintaining quality necessitates collaboration across roles, ensuring the article is well-structured, accurate, and engaging. 
To address these challenges, we propose WikiAutoGen framework. This framework facilitates structured collaboration among specialized agents, enabling comprehensive topic exploration, multi-modal content generation, and coherent generation. In the following sections, we provide a detailed breakdown of WikiAutoGen and its core components.

\subsection{WikiAutoGen Framework}

Our WikiAutoGen framework consists of several key components that work collaboratively to generate high-quality multimodal Wikipedia-style articles, as illustrated in Figure \ref{fig:pipeline}. Each module plays a distinct role in the article generation process:  1). \textbf{Outline Proposal Module.} This module converts the given text and image topics into structured outline proposals, laying the foundation for content organization.  
2). \textbf{Textual Article Writing Module.} This stage involves multiple sub-components, including a persona generator, a multi-agent discussion system, and the article generation process, ensuring the content is well-structured and contextually rich.  
3). \textbf{Multi-Perspective Self-reflection Module.} This component evaluates the generated text from multiple viewpoints, including those of a writer, reader, and editor, providing constructive feedback to refine and enhance the article.  
4). \textbf{Multimodal Article Writing Module.} This final stage integrates visual content, consisting of image positioning proposals, image retrieval, image selection, and multimodal refinement, ensuring a cohesive and well-balanced article presentation.  These components collectively enable WikiAutoGen to produce high-quality and multi-modal Wikipedia-style articles.

\subsection{Outline Proposal}

Given a multimodal topic, the first step is to interpret the input and generate topic-related outlines to guide knowledge exploration and information retrieval. We achieve it by leveraging LLMs~\cite{hurst2024gpt} and external search tools.

For text-based topics, the LLM analyzes the input, identifies relevant subtopics, and generates a structured outline to facilitate knowledge exploration. For image-based topics, we utilize Google Vision Search to retrieve metadata, including descriptions and contextual information. Named entity recognition (NER)~\cite{DBLP:journals/corr/abs-1810-04805} is then applied to extract the top 10 most frequent entities as query keywords. These keywords, combined with the original topic, are then fed into the LLM to generate a well-structured outline.
In the case of image-text topics, we combine insights from both modalities by extracting subtopics from the text and retrieving image metadata. The LLM then refines the topic and synthesizes a cohesive outline, integrating textual and visual information to guide comprehensive knowledge retrieval.

\subsection{Textual Article Writing}

The textual article writing process involves multiple components working together to generate a well-structured and informative article. It consists of a persona generator, multi-agent knowledge exploration, and article generation.

\noindent \textbf{Persona generator.} Given a draft outline, the LLM generates \(n\) distinct personas relevant to the topic (where n is a customizable parameter, $n \geq 1$),  each acting as an independent agent. The LLM assigns specific objectives to each agent based on their role. These agents are equipped with access to external web search tools
and contribute to a more comprehensive and well-supported article.

\noindent \textbf{Multi-agent knowledge exploration.} The knowledge exploration stage involves a fixed agent, the \textit{asker}, and n LLM-generated agents who are assigned specific roles. The \textit{asker} iterates through the outline, posing targeted questions, while the other agents search the internet for relevant information. They then share findings, discuss them, and refine their understanding.
During the discussion, they also interact with the multi-perspective self-reflection module, which provides feedback and improvement advice from a writer's perspective on \textit{reliability}, \textit{engagement}, \textit{consistency}, and \textit{informativeness}. This iterative process ensures a well-rounded knowledge base before article generation.

\noindent \textbf{Article generation.} 
After gathering web knowledge, the next step is to summarize the collected content and generate the textual article using an LLM-based writing agent. Once the initial draft is produced, the agent iteratively sends each generated section to the multi-perspective self-reflection module for feedback and refines each paragraph. This module evaluates the article from a writer’s perspective, providing suggestions to enhance coherence and readability. The writing agent incorporates these refinements, progressively improving the text to produce the final article.

\subsection{Multi-Perspective Self-reflection.} 
Writing a high-quality Wikipedia-style article requires addressing multiple aspects, including topic consistency, readability, engagement, and informativeness, to provide an optimal reading experience. Therefore, we introduce a multi-perspective self-reflection module that systematically evaluates and refines content across these dimensions. This module takes four distinct viewpoints and assesses the article from seven perspectives. 

\noindent \textbf{Perspectives.}
Our multi-perspective self-reflection focuses on the seven key criteria to improve the paper writing quality. They include \textit{reliability}, \textit{engagement}, \textit{informativeness}, \textit{coherence}, \textit{readability},  \textit{consistency} and \noindent \textit{helpfulness}. We provide a detailed explanation for them in the Appendix. 
  
\noindent \textbf{Supervisor viewpoint.} The supervisor assesses whether the generated content fully addresses the questions posed by the asker, evaluates the article’s depth, breadth, and coherence, and reviews the effectiveness of the multi-agent discussion. Additionally, it evaluates whether the generated content aligns with the topic and proposed outlines. Based on these criteria, the supervisor provides an evaluation and passes the feedback to the next role. Depending on the specific needs, the next role can be the \textit{writer}, \textit{reader}, or \textit{editor}.

\begin{table*}[t]
\vspace{-8mm}
\centering
\fontsize{8pt}{10pt}\selectfont
\resizebox{\textwidth}{!}{
\begin{tabular}{ c c c c c c c c c c c c}
\toprule
 & \multirow{3}{*}{\textbf{Methods}} & \multicolumn{4}{c}{\textbf{Content Quality}}  & \multicolumn{2}{c}{\textbf{Informativeness}} & \multicolumn{1}{c}{\textbf{Reliability}} &
 \multicolumn{2}{c}{\textbf{Engagement}} 
 & \multirow{2}{*}{\textbf{Average}}
\\
\cmidrule(lr){3-6} \cmidrule(lr){7-8} 
\cmidrule(lr){9-9} 
\cmidrule(lr){10-11} 
&  & {\textit{\textbf{Alignment}}}  & {\textit{\textbf{Consistency}}} & {\textit{\textbf{Relevance}}} & {\textit{\textbf{Repetition}}} & {\textit{\textbf{Breadth}}}  & {\textit{\textbf{Depth}}} & {\textit{\textbf{Verifiability}}}  & {\textit{\textbf{Engagement}}} & {\textit{\textbf{Novelty}}}\\
\cmidrule{1-12}
\multicolumn{12}{c}{\cellcolor{grayred} \textbf{\textit{Text as Topic}}} \\
\cmidrule{1-12}
& oRAG \cite{Asai2023SelfRAGLT}  &  61.35  & 73.96 & 76.04  & 71.11 & 63.30   & 52.42  & 45.47  & 57.51 & 45.24  & 60.71 \\
& Storm~\cite{Shao2024AssistingIW}  & 72.49  & 79.13 & 71.22  & 69.47 &  65.62  & 62.61 &  52.41 & 58.80  & 55.58  & 65.26   \\
& Co-Storm~\cite{Jiang2024IntoTU}  & 78.05 & 84.10 & 75.11  & 75.40  & 68.42   &  67.70 &  58.20 & 61.02  & 61.61  & 69.96   \\
& OmniThink~\cite{Xi2025OmniThinkEK} & 70.53 & 79.67   & 72.41  & 69.26  &  63.55 & 61.21 &  48.57 & 57.39  & 53.21 & 63.98  \\
\rowcolor{TealBlue}
& WikiAutoGen (Ours) & \textbf{81.68} & \textbf{90.87}  & \textbf{88.02}  & \textbf{83.62} & \textbf{79.64} &  \textbf{73.73}  & \textbf{70.69}  & \textbf{71.14}  & \textbf{69.21}  & \textbf{78.73}   \\
\cmidrule{1-12}
\multicolumn{12}{c}{\cellcolor{uclagold} \textbf{\textit{Image as Topic}}} \\
\cmidrule{1-12}
& oRAG & 50.10 & 72.16  &  50.92 & 65.47 & 42.01   & 43.26  & 33.90  & 40.66 & 36.91 &  48.38 \\
& Storm  & 45.80 & 59.60 &  45.99 & 46.38 &  42.69  &  39.92 & 34.23 & 42.38  & 35.17  & 43.57    \\
& Co-Storm  & 47.00 & 61.40 & 44.85  & 47.76 &  41.98 &  41.83  &  35.03 &  41.99 & 37.69 & 44.39   \\
& OmniThink & 43.61 & 58.29 & 43.03  & 45.67 & 38.63   & 42.26 &  29.18 & 38.31  & 33.57 & 41.39   \\
\rowcolor{TealBlue}
& WikiAutoGen (Ours) & \textbf{82.57} & \textbf{88.75} &  \textbf{87.20} & \textbf{80.22} &  \textbf{77.24}  & \textbf{74.99} & 
\textbf{68.41} &  \textbf{69.36} & \textbf{68.69} & \textbf{77.49}    \\
\cmidrule{1-12}
\multicolumn{12}{c}{\cellcolor{uclablue} \textbf{\textit{Image-Text as Topic}}} \\
\cmidrule{1-12}
& oRAG & 60.08 & 75.16  & 70.94  & 72.24 & 58.38   & 50.57  & 42.47  & 55.01 & 43.95 &  58.76 \\
& Storm  & 67.20 & 75.29  &  66.64 & 64.33 &  61.61 &  58.26 & 49.21 & 56.25  & 51.27  & 61.12   \\
& Co-Storm  & 70.15 & 79.29 & 67.31  & 68.89 &  61.90  & 61.22  & 52.44   & 57.05  & 55.68 & 63.77 \\
& OmniThink &  64.56 & 75.33 & 64.63  & 63.64  & 57.44   & 56.38 & 43.04 & 54.14  &  48.86 & 58.67  \\
\rowcolor{TealBlue}
& WikiAutoGen (Ours) &  \textbf{85.26} & \textbf{90.63} & \textbf{88.44}  & \textbf{82.11} &  \textbf{79.31}  & \textbf{75.20} & \textbf{68.59}  & \textbf{68.79}  & \textbf{71.06} & \textbf{78.82} \\
\bottomrule
\end{tabular}}
\vspace{2mm}
\caption{\textbf{Comparison of article generation performance for textual content.} We evaluate content quality, informativeness, reliability, and engagement under three input modalities (Text-only, Image-only, and Image-Text) on our WikiSeek benchmark.}
\vspace{-5mm}
\label{mainresult}
\end{table*}

\noindent \textbf{Writer viewpoint. } From the writer's viewpoint, the primary focus is on the multi-agent knowledge exploration and article generation stages. The writer evaluates whether the generated content maintains coherence, ensures engagement, verifies factual accuracy, and upholds logical consistency. Based on these assessments, the writer provides targeted improvement suggestions. 

For instance, to enhance coherence, it may be recommended to rearrange sentences or add transitional words and phrases. To improve readability, it might suggest simplifying complex concepts. Finally, the writer responds with a set of targeted and refined suggestions.

\noindent \textbf{Reader viewpoint. } To create a high-quality multi-modal article, it is essential to effectively integrate textual content with relevant images to enhance reader engagement and readability. To achieve this, our framework first employs an LLM to propose suitable image placements within the article and generate descriptive content specifying the types of images to include. This initial proposal is then reviewed by the multi-perspective self-reflection module, which evaluates the image positioning and the generated image descriptions from the perspective of readability, engagement, and helpfulness. Based on this assessment, the module provides constructive feedback, ensuring that visual content is effectively integrated to enrich the overall reader experience.

\noindent \textbf{Editor viewpoint.} After inserting images into the generated article, there may still be discrepancies or inconsistencies between the visual content and the corresponding textual descriptions. To address this, our framework sends the images along with their related text segments to the multi-perspective self-reflection module. This module evaluates the alignment and coherence between the images and their accompanying texts from an editorial viewpoint. It provides targeted suggestions, such as refining the image captions, adjusting image placement, or enhancing textual explanations to better reflect visual content. 
This final step ensures enhanced relevance, coherence, and readability between visual and textual elements.

\subsection{Multi-modal Article Writing}
Following textual generation, we incorporate relevant visual content to enhance the article’s readability and expressiveness.
This multimodal integration process involves several stages: image positioning proposal, image retrieval, image selection, and finally, an article refinement step that seamlessly integrates the images and text.

\noindent \textbf{Image positioning proposal.} 
After generating the complete textual article, an LLM-based agent is employed to propose appropriate placements and corresponding descriptions for relevant images.
These initial proposals are then refined through interaction with the multi-perspective self-reflection module, which evaluates their relevance, coherence, and engagement from the reader’s perspective.

\noindent \textbf{Image retrieval.} We then retrieve relevant images by performing searches based on multiple sources, including general image search engines, Wikipedia, and the websites mentioned in the references. After that, we can obtain a list of relevant image candidates.

\noindent \textbf{Image selection.} 
We first use the CLIP model~\cite{radford2021clip}  
o rank retrieved images based on semantic similarity to the generated captions, selecting the top-3 candidates.
Subsequently, we leverage a multi-modal model~\cite{hurst2024gpt} to further evaluate these candidates and select the most contextually appropriate image
for inclusion in the article.

\noindent \textbf{Article Polishing.} 
After finalizing image selection, we integrate the chosen images into the article and proceed to a polishing stage. Due to potential discrepancies between textual content and visual figures, we employ a multi-modal model to revise the entire article, enhancing coherence and consistency across modalities. Additionally, this multimodal model interacts with the multi-perspective self-reflection module, obtaining editorial feedback to further refine the integrated content.

\section{Experiment}
\label{experiment}
\begin{table*}[t]
\vspace{-10mm}
\centering
\fontsize{8pt}{10pt}\selectfont
\resizebox{\textwidth}{!}{
\begin{tabular}{c c c c c c c c c c c c c c}
\toprule

\multicolumn{3}{c}{\textbf{Modules}}   & 
\multicolumn{4}{c}{\textbf{Content Quality}}  & 
\multicolumn{2}{c}{\textbf{Informativeness}} & 
\multicolumn{1}{c}{\textbf{Reliability}} &
\multicolumn{2}{c}{\textbf{Engagement}} &
\multicolumn{2}{c}{\textbf{Average}} \\ 

\cmidrule(lr){1-3}  \cmidrule(lr){4-7} \cmidrule(lr){8-9} \cmidrule(lr){10-10} \cmidrule(lr){11-12} \cmidrule(lr){13-14}
{\textit{\textbf{Multi-agent}}} &  
\textit{\textbf{Outline proposal}} & 
{\textit{\textbf{Self-reflection}}}   &  
{\textit{\textbf{Alignment}}}  & 
{\textit{\textbf{Consistency}}} & 
{\textit{\textbf{Relevance}}} & 
{\textit{\textbf{Repetition}}} & 
{\textit{\textbf{Breadth}}}  & 
{\textit{\textbf{Depth}}} & 
{\textit{\textbf{Verifiability}}}  & 
{\textit{\textbf{Engagement}}} & 
{\textit{\textbf{Novelty}}} & 
\multicolumn{2}{c}{} \\ 

\cmidrule{1-14}
 \usym{1F5F6}  & \usym{1F5F6}   &   \usym{1F5F6}    & 50.63 & 62.87 & 53.10  & 51.19 & 48.58   & 44.35  & 43.64 & 45.54  &  39.08  & 48.78  \\
   \usym{2714}   & \usym{1F5F6}   &  \usym{1F5F6}   & 71.81 & 76.38  & 72.67  & 68.25 & 64.36  &  69.16 & 64.20  & 57.04  &  55.17  &  66.56 \\
 \usym{1F5F6}  &  \usym{2714}   &  \usym{1F5F6}    & 79.11 & 86.20 & 80.93  & 77.21  &  72.65 & 62.13  & \textbf{69.19}  & 64.45  & 64.11  & 72.88 \\
 \usym{1F5F6}   &  \usym{1F5F6}   & \usym{2714}  & 75.91  & 84.08 &  82.20 & 75.56 & 73.24  & 66.66  & 65.41 &  63.03 & 58.36    &  71.60 \\
   \usym{1F5F6}   & \usym{2714}  &  \usym{2714}  & 77.68  &  85.57 & 84.73 &   78.49 & 75.08  &  68.85  & 65.59  & 65.73  & 66.45  & 73.80  \\
\rowcolor{TealBlue}
\usym{2714}  &  \usym{2714}  & \usym{2714}  &  \textbf{82.57} & \textbf{88.75} & \textbf{87.20}  & \textbf{80.22} & \textbf{77.24}  & \textbf{74.99}   & 68.41  & \textbf{69.36}  & \textbf{68.69}  & \textbf{77.49}  \\
\bottomrule
\end{tabular}
}
\vspace{-2mm}
\caption{ \textbf{Ablation study.} We study the impact of individual modules (Multi-agent knowledge exploration, outline proposal, and self-reflection) on article generation performance for text content. The input modality is image-only. }
\vspace{-4mm}
\label{ablation_study1}
\end{table*}

\begin{table}[t]
\centering
\fontsize{10pt}{10pt}\selectfont
\resizebox{\linewidth}{!}{%
\begin{tabular}{c c c c c c c}
\toprule
 & {\textit{\textbf{Method}}} & {\textit{\textbf{Coherence}}} & {\textit{\textbf{Engagement}}} & {\textit{\textbf{Helpfulness}}}  & {\textit{\textbf{Info. Sup.}}}   & {\textit{\textbf{Average}}}  \\
\cmidrule{1-7}
\multicolumn{7}{c}{\cellcolor{grayred} \textbf{\textit{Text as Topic}}} \\
\cmidrule{1-7}
 & oRAG & 57.36 & 56.26 &  63.61 & 51.90 &  57.28     \\ 
 & Storm  &  55.20 & 45.97 & 51.89 & 43.97   &  49.26  \\ 
 & Co-Storm   & 57.62  & 48.64 & 54.19 & 45.07  &  51.38  \\
 & OmniThink  & 58.82  & 49.36 & 54.93 & 47.55   &  52.67  \\
 \rowcolor{TealBlue}
 & WikiAutoGen (Ours) & \textbf{70.12}  & \textbf{66.31} & \textbf{74.76} & \textbf{64.78}  &  \textbf{68.99}  \\
\cmidrule{1-7}
\multicolumn{7}{c}{\cellcolor{uclagold} \textbf{\textit{Image as Topic}}} \\
\cmidrule{1-7}
 & oRAG & 61.21& 52.07 & 58.039 & 45.96 & 54.32       \\  
 & Storm  & 52.59 & 43.98 & 49.53 &  41.84  &  46.99  \\ 
 & Co-Storm   & 54.32 &  45.61  & 51.55 & 41.56 & 48.26 \\
 & OmniThink  &  59.88 & 50.62 & 56.13 & 47.23 & 53.47    \\ 
 \rowcolor{TealBlue}
 & WikiAutoGen (Ours) & \textbf{71.90}  & \textbf{61.69} & \textbf{77.63} & \textbf{63.88} &  \textbf{68.78}  \\
\cmidrule{1-7}
\multicolumn{7}{c}{\cellcolor{uclablue} \textbf{\textit{Image-Text as Topic}}} \\
\cmidrule{1-7}
 & oRAG & 66.38 & 56.31 & 62.94 & 49.78  & 58.85    \\ 
 & Storm  & 59.67 & 50.26 & 55.78 & 46.78  &  53.12  \\ 
 & Co-Storm   & 54.28 & 45.65  &51.29  & 42.88  &  48.53   \\ 
 & OmniThink  & 61.76  & 51.40 & 57.88 &  49.97  &   55.25  \\ 
 \rowcolor{TealBlue}
 & WikiAutoGen (Ours) &\textbf{72.24}  & \textbf{70.29} & \textbf{72.11} & \textbf{69.29}  & \textbf{70.98} \\
\bottomrule
\end{tabular}%
}
\caption{\textbf{Comparison of article generation performance for image content.} We evaluate textual generation for image-text coherence, image engagement, image helpfulness, and information supplement on our WikiSeek benchmark.}
\label{vision_results}
\vspace{-5mm}
\end{table}

\subsection{Experiment setup}

\noindent \textbf{Implementation Details.} 
For the language model (LM) components of WikiAutoGen, we employ zero-shot prompting implemented using the DSPy framework~\cite{khattab2024dspy} in conjunction with GPT-4o~\cite{hurst2024gpt}\footnote{All models and datasets used in this work were accessed solely by authors at KAUST.}, GPT-4o-mini, and GPT-o3-mini~\cite{openai2025o3mini}. Specifically, we use GPT-o3-mini for the multi-perspective self-reflection module due to its strong reasoning capabilities, GPT-4o for the multimodal knowledge exploration tasks, and GPT-4o-mini for all other remaining tasks. 
WikiAutoGen retrieves real-time web information via Serper’s API\footnote{https://serper.dev/}, with each query returning up to 5 web pages. Throughout the experiments, we maintain consistent settings by fixing the LM temperature at 1.0 and the $top\_p$ value at 0.9.
For evaluation, we utilize GPT-4o as the evaluator. To further validate the results, we conduct more evaluations using two distinct
evaluators, Gemini2.5-Flash~\cite{gemini} and Prometheus2~\cite{kim2024prometheus}, as detailed in Appendix~B.

\noindent \textbf{Evaluation metrics.}
We evaluate the generated multimodal articles through separate assessments of their textual and visual content.
 
\textbf{- Text quality evaluation.} 
Following prior evaluation frameworks~\cite{Shao2024AssistingIW, Jiang2024IntoTU}, we utilize GPT-4o as the evaluator to assess generated articles across nine criteria grouped into four main aspects: 

\begin{itemize}
    \item \textit{Content quality}: alignment, relevance, repetition, and consistency;
    \item \textit{Informativeness}: breadth and depth;
    \item \textit{Reliability}: verifiability;
    \item \textit{Engagement}: engagement and novelty.
\end{itemize}

\textbf{- Image quality evaluation.} As there are no existing benchmarks specifically designed for evaluating multimodal Wikipedia-style article generation, we propose an evaluation method inspired by the textual evaluation frameworks~\cite{Shao2024AssistingIW, Jiang2024IntoTU} to evaluate image quality. Specifically, we assess image quality based on four criteria: \textit{image-text coherence}, \textit{engagement}, \textit{helpfulness}, and \textit{information supplement} (the image's ability to provide additional useful information beyond the textual context).

\noindent \textbf{Baselines}
We compare WikiAutoGen with four representative LLM-based baseline frameworks for automated expository writing on our WikiSeek benchmark:

\begin{itemize}
    \item \textbf{Outline-driven RAG (oRAG)} generates articles guided by outlines produced by Self-RAG~\cite{Asai2023SelfRAGLT}.
    \item \textbf{Storm}~\cite{Shao2024AssistingIW} leverages LLM-driven conversations and outlines from diverse perspectives to generate Wikipedia-style content.
    \item \textbf{Co-STORM}~\cite{Jiang2024IntoTU} utilizes collaborative discourse among multiple LLM agents to explore and discover unknown information.
    \item \textbf{OmniThink}~\cite{Xi2025OmniThinkEK} enhances article quality through iterative expansion and reflection, emulating human slow-thinking to increase knowledge density.
\end{itemize}

Since these baselines originally lack multimodal capabilities, we equip them with image retrieval functionalities to enable a fair multimodal comparison. Specifically, each baseline can also retrieve images via Google image search, extract relevant metadata, and search for images based on generated textual descriptions.

\subsection{Experimental results}

\noindent \textbf{Textual content comparison on WikiSeek.}
We demonstrate our results in the Table \ref{mainresult}. The results indicate that WikiAutoGen consistently achieves the highest average scores across all evaluation dimensions—content quality, informativeness, reliability, and engagement—highlighting its comprehensive effectiveness in article generation.

For text-only inputs, our WikiAutoGen achieves an average score of 78.73, significantly outperforming the best baseline (Co-Storm, 69.96) by approximately 8.8 points, demonstrating superior coherence, alignment, and informativeness. In image-only scenarios, the improvement is even more pronounced, with WikiAutoGen obtaining 77.49, surpassing the strongest baseline (oRAG, 48.38) by approximately 29.1 points, reflecting its exceptional capability to extract meaningful textual insights from visual content. For combined image-text topics, our model maintains its advantage with an average score of 78.82, showing a clear improvement (+ 15.05 points) over the next-best baseline (Co-Storm, 63.77). Overall, the substantial performance gains demonstrate that WikiAutoGen excels at synthesizing cohesive and engaging content across diverse inputs.

\noindent \textbf{Image content comparison on WikiSeek.} Table \ref{vision_results} compares the performance of image content across different article generation methods. Our WikiAutoGen consistently achieves the highest scores across all image evaluation metrics for all three input modalities. 

Specifically, our method significantly improves upon baseline methods, outperforming the next-best method by approximately 11.34 points on image-text coherence (72.24 vs. Storm’s 59.67) under the image-text topics. Similarly, for image-only topics, our approach excels particularly in helpfulness (77.63) and coherence (71.90), demonstrating WikiAutoGen's superior capability in selecting images that meaningfully complement textual content and provide additional useful information. 
Overall, these results highlight WikiAutoGen's effectiveness in integrating images to enhance coherence, engagement, and informativeness, substantially advancing the overall quality and readability of multimodal articles.

\subsection{Ablation studies}

\begin{figure}[t]
    \centering
    \vspace{-2mm}
    \includegraphics[width=1\linewidth]{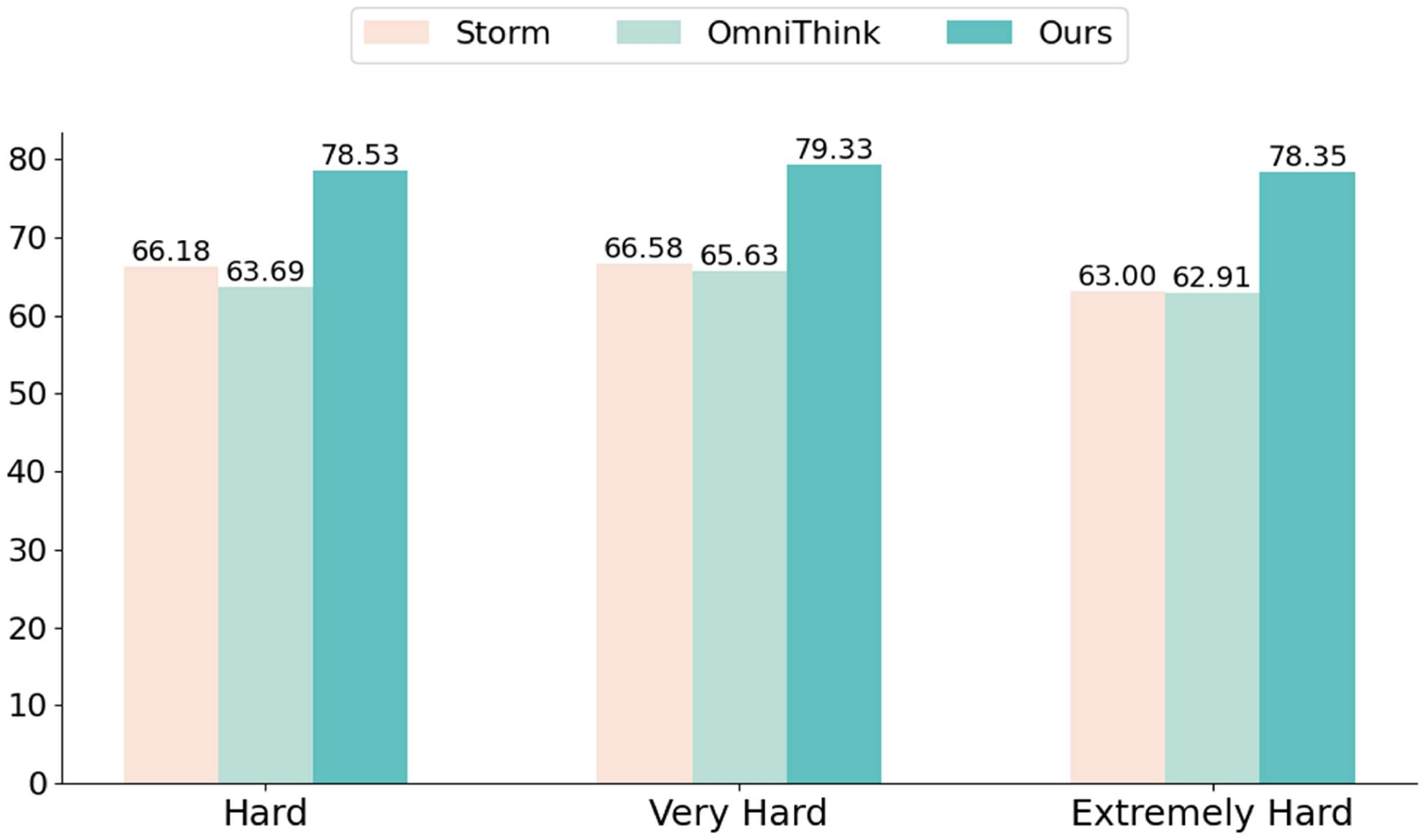}
    \vspace{-5mm}
    \caption{\textbf{Ablation study across different data difficulty levels.} We compare our WikiAutoGen with Storm and OmniThink on three difficulty categories: hard (300–500 characters), very hard (100–300 characters), and extremely hard ($<$100 characters).}
    \label{fig:Quantitative Comparisons}
    \vspace{-5mm}
\end{figure}

\noindent \textbf{Ablation on different components of WikiAutoGen.} 
We conduct an ablation study to analyze the individual contributions of three core components to textual article generation from image-only topics. 
Specifically, we examine the impact of components, including multi-agent knowledge exploration, outline proposal, and multi-perspective self-reflection.
Results are shown in Table~\ref{ablation_study1}. Specifically, without the outline proposal, the system simply retrieves a single image and extracts its description from metadata. Without multi-agent knowledge exploration, only a single static agent responds to the asker.

Incorporating the multi-agent module improves performance from 48.78 (baseline without modules) to 66.56 (+17.78 points), 
highlighting its effectiveness in collaborative knowledge exploration. Using the outline proposal alone further increases performance to 72.88 (+24.10 points), underscoring its importance for content structuring. The self-reflection module individually achieves 71.60 (+22.82 points), indicating its strength in refining coherence and consistency. Combining all three modules results in the highest performance (77.49), validating their complementary roles in generating coherent, informative, and engaging articles from image-only inputs.

\noindent \textbf{Ablation on different difficulty levels.} 
In our WikiSeek benchmark, topics are grouped into three difficulty levels based on article length (character count): hard (300–500 characters), very hard (100–300 characters), and extremely hard (fewer than 100 characters), with 100 examples per level.
We evaluate text-only inputs and present the average textual evaluation results in Figure~\ref{fig:Quantitative Comparisons}.
The results indicate that our WikiAutoGen consistently outperforms all baseline methods across all three difficulty levels. Notably, as the topic difficulty increases (from hard to extremely hard), the performance gap between WikiAutoGen and Storm widens from 12.35 points to 15.35 points, with similar trends observed against other baselines. These results demonstrate the robustness and stability of WikiAutoGen in effectively handling highly challenging and underexplored topics.

\subsection{Compared with Commercial Deep Research}

\begin{table}[h!]
    \centering
    \vspace{-3.5mm}
    \resizebox{\linewidth}{!}{
      \begin{tabular}{cccccc}
        \toprule
        \textbf{Method}
        & \textbf{Text as Topic} 
        & \textbf{Image as Topic}
        & \textbf{Image-Text as Topic} 
        & \textbf{Average} 
        & \textbf{Compute Time} \\
        \midrule
        OpenAI      & 92.75 & 91.95 & 94.50 & 93.06 & $\sim\,30\ \mathrm{min}$ \\
        Google      & 81.91 & \textemdash{} & \textemdash{} & \textemdash{} & $\sim\,12\ \mathrm{min}$ \\
        Grok        & 88.35 & 81.70 & 86.13 & 85.06 & $\sim\,10\ \mathrm{min}$ \\
        \rowcolor{TealBlue}
        WikiAutoGen & 88.58 & 89.55 & 88.89 & 89.01 & $\sim\,\mathbf{8}\ \mathrm{min}$ \\
        \bottomrule
      \end{tabular}
    }
    \caption{Comparison of commercial models' article generation performance with Prometheus2~\cite{kim2024prometheus}.}
    \vspace{-4.5mm}
    \label{tab:deep}
\end{table}

We evaluate the performance of WikiAutoGen against leading commercial models on the WikiSeek benchmark, spanning three input modalities. As shown in Table~\ref{tab:deep}, WikiAutoGen achieves an impressive average score of 89.01, closely approaching OpenAI (93.06), while significantly outperforming Grok (85.06) and Google (81.91). WikiAutoGen delivers consistently strong results across text (88.58), image (89.55), and image-text (88.89) inputs, demonstrating robust cross-modal capabilities.
In terms of efficiency, WikiAutoGen generates articles in just 8 minutes, making it over 3.75$\times$ faster than OpenAI (30 minutes), the fastest among the commercial baselines. This substantial speed advantage underscores its practical scalability and suitability for real-world, time-sensitive applications.

\subsection{Human evaluation}
\begin{figure}[h]
    \centering
    \vspace{-2mm}
    \includegraphics[width=1\linewidth]{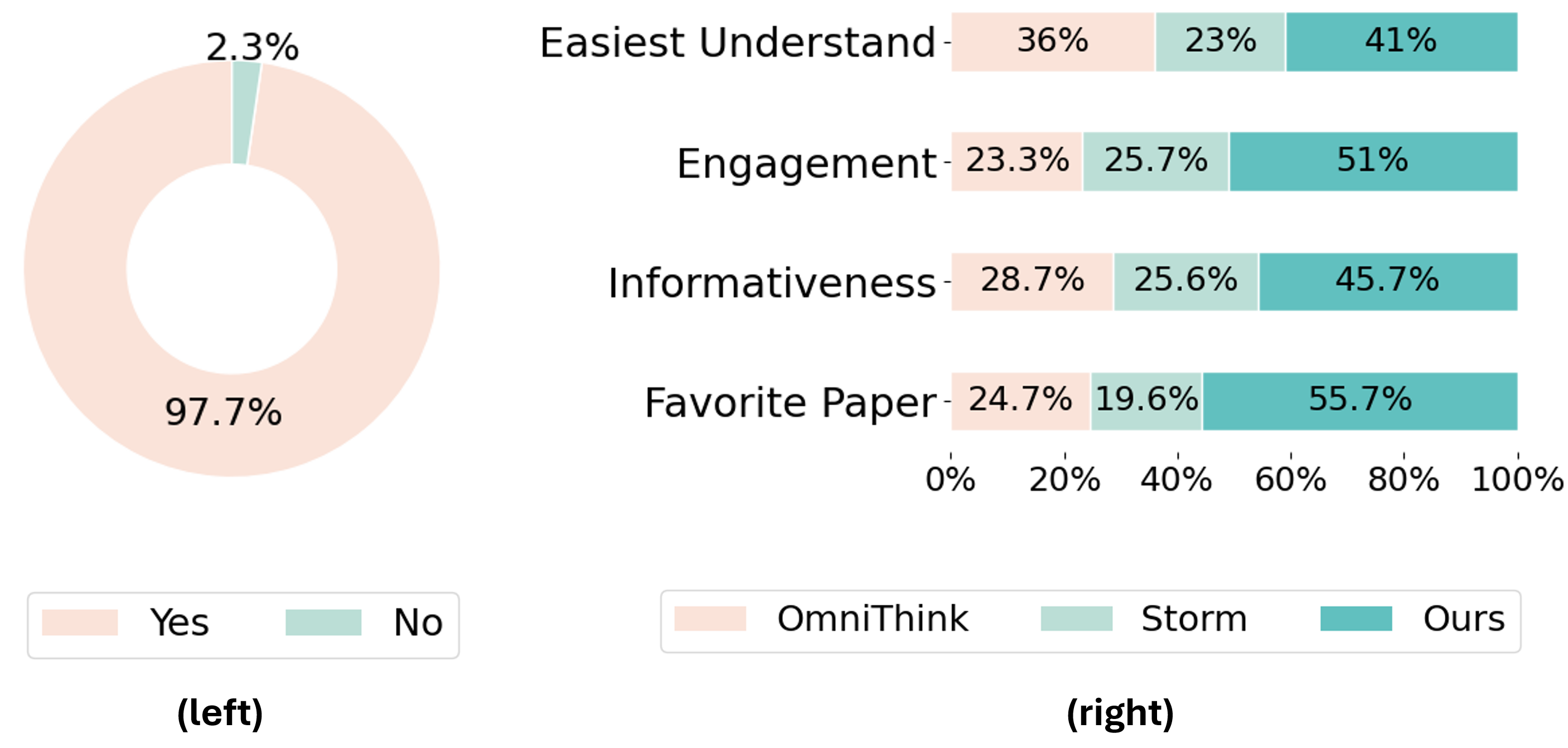}
    \vspace{-6mm}
    \caption{\textbf{Human evaluation study.} We randomly sample text-only topics and conduct comparative evaluations between WikiAutoGen, Storm, and OmniThink. \textbf{Left:} Participants respond to the question, \textit{``Do you think adding images would improve comprehension of the topic?"}. \textbf{Right:} Participants answer multiple-choice questions evaluating readability, engagement, informativeness, and overall preference.}
    \label{fig:HumanEvaluation}
    \vspace{-5mm}
\end{figure}

To evaluate the quality of the generated Wikipedia-style articles, we conduct a human evaluation study via Amazon Mechanical Turk (AMT)\footnote{\href{https://www.mturk.com/}{https://www.mturk.com/}}. We randomly sample 100 text-only topics from the WikiSeek benchmark dataset
and perform pairwise comparisons between articles generated by our method (WikiAutoGen), Storm, and OmniThink.
Each topic is evaluated by three independent participants in randomized order.
Participants first answer the question: \emph{``Do you think adding images would improve comprehension of the topic?''} 
As shown in Figure~\ref{fig:HumanEvaluation} (left), 97.7\% of participants agree that images improve topic comprehension.

Additionally, participants answer multiple-choice questions, including: (1) Which article is the easiest to understand? (2) Which article is the most engaging in terms of narrative, examples, or overall presentation? (3) Which article provides the most comprehensive background information and in-depth analysis?  (4) Which article is your overall favorite? 
The question order is randomized to mitigate potential evaluation bias.
As illustrated in Figure~\ref{fig:HumanEvaluation} (right), participants consistently prefer articles generated by WikiAutoGen over those by Storm and OmniThink across all evaluation criteria.

\vspace{-1mm}
\section{Conclusion}
\label{con}
In this paper, we introduced WikiAutoGen, a comprehensive multi-agent framework designed for automated multimodal Wikipedia-style article generation. WikiAutoGen integrates both visual and textual content, significantly enhancing the depth, informativeness, and engagement of generated articles. 
To address key limitations in prior work, we proposed a novel multi-perspective self-reflection mechanism, which systematically improves article coherence, reliability, and overall quality.
Furthermore, we presented WikiSeek, a challenging multimodal benchmark specifically crafted to evaluate the performance of models in generating content for less-explored topics. Experimental results demonstrated that WikiAutoGen substantially outperforms existing state-of-the-art baselines across both textual and visual evaluation metrics.
Particularly notable were the improvements in content quality, informativeness, and reader engagement, validating the effectiveness of integrating multimodal inputs and iterative self-reflection.

{
    \small
    \bibliographystyle{ieeenat_fullname}
    \bibliography{main}

\begin{thebibliography}{45}
\providecommand{\natexlab}[1]{#1}
\providecommand{\url}[1]{\texttt{#1}}
\expandafter\ifx\csname urlstyle\endcsname\relax
  \providecommand{\doi}[1]{doi: #1}\else
  \providecommand{\doi}{doi: \begingroup \urlstyle{rm}\Url}\fi

\bibitem[Asai et~al.(2023)Asai, Wu, Wang, Sil, and Hajishirzi]{Asai2023SelfRAGLT}
Akari Asai, Zeqiu Wu, Yizhong Wang, Avirup Sil, and Hannaneh Hajishirzi.
\newblock Self-rag: Learning to retrieve, generate, and critique through self-reflection.
\newblock \emph{ArXiv}, abs/2310.11511, 2023.

\bibitem[Bai et~al.(2023)]{qwen}
Jinze Bai et~al.
\newblock Qwen technical report.
\newblock \emph{arXiv preprint arXiv:2309.16609}, 2023.
\newblock Accessed: 2025-03-04.

\bibitem[Balepur et~al.(2023)Balepur, Huang, and Chang]{Balepur2023ExpositoryTG}
Nishant Balepur, Jie Huang, and Kevin Chen-Chuan Chang.
\newblock Expository text generation: Imitate, retrieve, paraphrase.
\newblock \emph{ArXiv}, abs/2305.03276, 2023.

\bibitem[Besta et~al.(2024)Besta, Memedi, Zhang, Gerstenberger, Piao, Blach, Nyczyk, Copik, Kwa'sniewski, Muller, Gianinazzi, Kub\v{c}ek, Niewiadomski, O'Mahony, Mutlu, and Hoefler]{Besta2024DemystifyingCT}
Maciej Besta, Florim Memedi, Zhenyu Zhang, Robert Gerstenberger, Guangyuan Piao, Nils Blach, Piotr Nyczyk, Marcin Copik, Grzegorz Kwa'sniewski, Jurgen Muller, Lukas Gianinazzi, Ale\v{s} Kub\v{c}ek, Hubert Niewiadomski, Aidan O'Mahony, Onur Mutlu, and Torsten Hoefler.
\newblock Demystifying chains, trees, and graphs of thoughts.
\newblock In \emph{ArXiv}, 2024.

\bibitem[Bi et~al.(2024)]{deepseek}
DeepSeek-AI:~Xiao Bi et~al.
\newblock Deepseek llm: Scaling open-source language models with longtermism.
\newblock \emph{arXiv preprint arXiv:2401.02954}, 2024.
\newblock Accessed: 2025-03-04.

\bibitem[Burns et~al.(2023)Burns, Srinivasan, Ainslie, Brown, Plummer, Saenko, Ni, and Guo]{Burns2023WikiWeb2MAP}
Andrea Burns, Krishna Srinivasan, Joshua Ainslie, Geoff Brown, Bryan~A. Plummer, Kate Saenko, Jianmo Ni, and Mandy Guo.
\newblock Wikiweb2m: A page-level multimodal wikipedia dataset.
\newblock \emph{ArXiv}, abs/2305.05432, 2023.

\bibitem[Coyne(2023)]{coyne-2023-template}
Steven Coyne.
\newblock Template-guided grammatical error feedback comment generation.
\newblock In \emph{Proceedings of the 17th Conference of the European Chapter of the Association for Computational Linguistics: Student Research Workshop}, pages 94--104, Dubrovnik, Croatia, 2023. Association for Computational Linguistics.

\bibitem[DeepMind(2023)]{gemini}
Google DeepMind.
\newblock Gemini: Multimodal language model.
\newblock \url{https://deepmind.google/technologies/gemini/}, 2023.
\newblock Accessed: 2025-03-04.

\bibitem[Devlin et~al.(2018)Devlin, Chang, Lee, and Toutanova]{DBLP:journals/corr/abs-1810-04805}
Jacob Devlin, Ming{-}Wei Chang, Kenton Lee, and Kristina Toutanova.
\newblock {BERT:} pre-training of deep bidirectional transformers for language understanding.
\newblock \emph{CoRR}, abs/1810.04805, 2018.

\bibitem[Dong et~al.(2024)Dong, Tang, Li, Zhao, and Wen]{dong-etal-2024-bamboo}
Zican Dong, Tianyi Tang, Junyi Li, Wayne~Xin Zhao, and Ji-Rong Wen.
\newblock {BAMBOO}: A comprehensive benchmark for evaluating long text modeling capacities of large language models.
\newblock In \emph{Proceedings of the 2024 Joint International Conference on Computational Linguistics, Language Resources and Evaluation (LREC-COLING 2024)}, pages 2086--2099, Torino, Italia, 2024. ELRA and ICCL.

\bibitem[Gao et~al.(2023)Gao, Jiang, Yang, Zeng, Lu, Blum, Liu, She, Jiang, and Li]{Gao2023EvaluatingLL}
Fan Gao, Hang Jiang, Rui Yang, Qingcheng Zeng, Jinghui Lu, Moritz Blum, Dairui Liu, Tianwei She, Yuang Jiang, and Irene Li.
\newblock Evaluating large language models on wikipedia-style survey generation.
\newblock In \emph{Annual Meeting of the Association for Computational Linguistics}, 2023.

\bibitem[Gou et~al.(2023)Gou, Shao, Gong, Shen, Yang, Duan, and Chen]{Gou2023CRITICLL}
Zhibin Gou, Zhihong Shao, Yeyun Gong, Yelong Shen, Yujiu Yang, Nan Duan, and Weizhu Chen.
\newblock Critic: Large language models can self-correct with tool-interactive critiquing.
\newblock \emph{ArXiv}, abs/2305.11738, 2023.

\bibitem[Hurst et~al.(2024)Hurst, Lerer, Goucher, Perelman, Ramesh, Clark, Ostrow, Welihinda, Hayes, Radford, et~al.]{hurst2024gpt}
Aaron Hurst, Adam Lerer, Adam~P Goucher, Adam Perelman, Aditya Ramesh, Aidan Clark, AJ Ostrow, Akila Welihinda, Alan Hayes, Alec Radford, et~al.
\newblock Gpt-4o system card.
\newblock \emph{arXiv preprint arXiv:2410.21276}, 2024.

\bibitem[Jiang et~al.(2024)Jiang, Shao, Ma, Semnani, and Lam]{Jiang2024IntoTU}
Yucheng Jiang, Yijia Shao, Dekun Ma, Sina~J. Semnani, and Monica~S. Lam.
\newblock Into the unknown unknowns: Engaged human learning through participation in language model agent conversations.
\newblock In \emph{Conference on Empirical Methods in Natural Language Processing}, 2024.

\bibitem[Khattab et~al.(2024)Khattab, Singhvi, Maheshwari, Zhang, Santhanam, Vardhamanan, Haq, Sharma, Joshi, Moazam, Miller, Zaharia, and Potts]{khattab2024dspy}
Omar Khattab, Arnav Singhvi, Paridhi Maheshwari, Zhiyuan Zhang, Keshav Santhanam, Sri Vardhamanan, Saiful Haq, Ashutosh Sharma, Thomas~T. Joshi, Hanna Moazam, Heather Miller, Matei Zaharia, and Christopher Potts.
\newblock {DSPy}: Compiling declarative language model calls into self-improving pipelines.
\newblock In \emph{Proceedings of the Twelfth International Conference on Learning Representations (ICLR)}, 2024.

\bibitem[Kim et~al.(2024)Kim, Suk, Longpre, Lin, Shin, Welleck, Neubig, Lee, Lee, and Seo]{kim2024prometheus}
Seungone Kim, Juyoung Suk, Shayne Longpre, Bill~Yuchen Lin, Jamin Shin, Sean Welleck, Graham Neubig, Moontae Lee, Kyungjae Lee, and Minjoon Seo.
\newblock Prometheus 2: An open source language model specialized in evaluating other language models.
\newblock \emph{arXiv preprint arXiv:2405.01535}, 2024.

\bibitem[Kumar et~al.(2024)Kumar, Viswanathan, Yerra, Salemi, Rossi, Dernoncourt, Deilamsalehy, Chen, Zhang, Agarwal, et~al.]{kumar2024longlamp}
Ishita Kumar, Snigdha Viswanathan, Sushrita Yerra, Alireza Salemi, Ryan~A Rossi, Franck Dernoncourt, Hanieh Deilamsalehy, Xiang Chen, Ruiyi Zhang, Shubham Agarwal, et~al.
\newblock Longlamp: A benchmark for personalized long-form text generation.
\newblock \emph{arXiv preprint arXiv:2407.11016}, 2024.

\bibitem[Li et~al.(2021)Li, Fabbri, Kawamura, Liu, Tang, Tae, Shen, Ma, Mizutani, and Radev]{Li2021Surfer100GS}
Irene Li, Alexander~R. Fabbri, Rina Kawamura, Yixin Liu, Xiangru Tang, Jaesung Tae, Chang Shen, Sally Ma, Tomoe Mizutani, and Dragomir~R. Radev.
\newblock Surfer100: Generating surveys from web resources, wikipedia-style.
\newblock In \emph{International Conference on Language Resources and Evaluation}, 2021.

\bibitem[Liang et~al.(2024)Liang, Wu, Zhuang, Chen, Shen, Jia, Qin, Sanghai, Wang, Yang, and Bendersky]{Liang2024IntegratingPI}
Yi Liang, You Wu, Honglei Zhuang, Li Chen, Jiaming Shen, Yiling Jia, Zhen Qin, Sumit~K. Sanghai, Xuanhui Wang, Carl Yang, and Michael Bendersky.
\newblock Integrating planning into single-turn long-form text generation.
\newblock \emph{ArXiv}, abs/2410.06203, 2024.

\bibitem[Madaan et~al.(2023)Madaan, Tandon, Gupta, Hallinan, Gao, Wiegreffe, Alon, Dziri, Prabhumoye, Yang, Welleck, Majumder, Gupta, Yazdanbakhsh, and Clark]{Madaan2023SelfRefineIR}
Aman Madaan, Niket Tandon, Prakhar Gupta, Skyler Hallinan, Luyu Gao, Sarah Wiegreffe, Uri Alon, Nouha Dziri, Shrimai Prabhumoye, Yiming Yang, Sean Welleck, Bodhisattwa~Prasad Majumder, Shashank Gupta, Amir Yazdanbakhsh, and Peter Clark.
\newblock Self-refine: Iterative refinement with self-feedback.
\newblock \emph{ArXiv}, abs/2303.17651, 2023.

\bibitem[Min et~al.(2023)Min, Krishna, Lyu, Lewis, Yih, Koh, Iyyer, Zettlemoyer, and Hajishirzi]{min-etal-2023-factscore}
Sewon Min, Kalpesh Krishna, Xinxi Lyu, Mike Lewis, Wen-tau Yih, Pang Koh, Mohit Iyyer, Luke Zettlemoyer, and Hannaneh Hajishirzi.
\newblock {FA}ct{S}core: Fine-grained atomic evaluation of factual precision in long form text generation.
\newblock In \emph{Proceedings of the 2023 Conference on Empirical Methods in Natural Language Processing}, pages 12076--12100, Singapore, 2023. Association for Computational Linguistics.

\bibitem[OpenAI(2022)]{chatgpt}
OpenAI.
\newblock Chatgpt: Optimizing language models for dialogue.
\newblock \url{https://openai.com/blog/chatgpt}, 2022.
\newblock Accessed: 2025-03-04.

\bibitem[OpenAI(2025)]{openai2025o3mini}
OpenAI.
\newblock Openai o3-mini, 2025.

\bibitem[Paul et~al.(2023)Paul, Ismayilzada, Peyrard, Borges, Bosselut, West, and Faltings]{Paul2023REFINERRF}
Debjit Paul, Mete Ismayilzada, Maxime Peyrard, Beatriz Borges, Antoine Bosselut, Robert West, and Boi Faltings.
\newblock Refiner: Reasoning feedback on intermediate representations.
\newblock In \emph{Conference of the European Chapter of the Association for Computational Linguistics}, 2023.

\bibitem[Pich{\'e} et~al.(2024)Pich{\'e}, Milios, Bahdanau, and Pal]{Pich2024LLMsCL}
Alexandre Pich{\'e}, Aristides Milios, Dzmitry Bahdanau, and Chris Pal.
\newblock Llms can learn self-restraint through iterative self-reflection.
\newblock \emph{ArXiv}, abs/2405.13022, 2024.

\bibitem[Que et~al.(2024)Que, Duan, He, Mou, Zhou, Liu, Rong, Wang, Yang, Zhang, et~al.]{que2024hellobench}
Haoran Que, Feiyu Duan, Liqun He, Yutao Mou, Wangchunshu Zhou, Jiaheng Liu, Wenge Rong, Zekun~Moore Wang, Jian Yang, Ge Zhang, et~al.
\newblock Hellobench: Evaluating long text generation capabilities of large language models.
\newblock \emph{arXiv preprint arXiv:2409.16191}, 2024.

\bibitem[Radford et~al.(2021)Radford, Kim, Hallacy, Ramesh, Goh, Agarwal, Sastry, Askell, Mishkin, Clark, Krueger, and Sutskever]{radford2021clip}
Alec Radford, Jong~Wook Kim, Chris Hallacy, Aditya Ramesh, Gabriel Goh, Sandhini Agarwal, Girish Sastry, Amanda Askell, Pamela Mishkin, Jack Clark, Gretchen Krueger, and Ilya Sutskever.
\newblock Learning transferable visual models from natural language supervision.
\newblock In \emph{Proceedings of the 38th International Conference on Machine Learning (ICML)}, pages 8748--8763. PMLR, 2021.

\bibitem[Shao et~al.(2024)Shao, Jiang, Kanell, Xu, Khattab, and Lam]{Shao2024AssistingIW}
Yijia Shao, Yucheng Jiang, Theodore~A. Kanell, Peter Xu, Omar Khattab, and Monica~S. Lam.
\newblock Assisting in writing wikipedia-like articles from scratch with large language models.
\newblock In \emph{North American Chapter of the Association for Computational Linguistics}, 2024.

\bibitem[Shen et~al.(2023)Shen, August, Siangliulue, Lo, Bragg, Hammerbacher, Downey, and Chang]{Shen2023BeyondSD}
Zejiang Shen, Tal August, Pao Siangliulue, Kyle Lo, Jonathan Bragg, Jeff Hammerbacher, Doug Downey, and Joseph~Chee Chang.
\newblock Beyond summarization: Designing ai support for real-world expository writing tasks.
\newblock \emph{ArXiv}, abs/2304.02623, 2023.

\bibitem[Shinn et~al.(2023)Shinn, Cassano, Labash, Gopinath, Narasimhan, and Yao]{Shinn2023ReflexionLA}
Noah Shinn, Federico Cassano, Beck Labash, Ashwin Gopinath, Karthik Narasimhan, and Shunyu Yao.
\newblock Reflexion: language agents with verbal reinforcement learning.
\newblock In \emph{Neural Information Processing Systems}, 2023.

\bibitem[Sun et~al.(2023)Sun, Shen, Zhou, Zhang, Chen, Cox, Yang, and Gan]{Sun2023PrincipleDrivenSO}
Zhiqing Sun, Yikang Shen, Qinhong Zhou, Hongxin Zhang, Zhenfang Chen, David~D. Cox, Yiming Yang, and Chuang Gan.
\newblock Principle-driven self-alignment of language models from scratch with minimal human supervision.
\newblock \emph{ArXiv}, abs/2305.03047, 2023.

\bibitem[Tan et~al.(2021)Tan, Yang, Al-Shedivat, Xing, and Hu]{tan-etal-2021-progressive}
Bowen Tan, Zichao Yang, Maruan Al-Shedivat, Eric Xing, and Zhiting Hu.
\newblock Progressive generation of long text with pretrained language models.
\newblock In \emph{Proceedings of the 2021 Conference of the North American Chapter of the Association for Computational Linguistics: Human Language Technologies}, pages 4313--4324, Online, 2021. Association for Computational Linguistics.

\bibitem[Tan et~al.(2024)Tan, Guo, Shi, Xu, Liu, Feng, Li, Wang, Shang, Liu, et~al.]{tan2024proxyqa}
Haochen Tan, Zhijiang Guo, Zhan Shi, Lu Xu, Zhili Liu, Yunlong Feng, Xiaoguang Li, Yasheng Wang, Lifeng Shang, Qun Liu, et~al.
\newblock Proxyqa: An alternative framework for evaluating long-form text generation with large language models.
\newblock \emph{arXiv preprint arXiv:2401.15042}, 2024.

\bibitem[Tang et~al.(2023)Tang, Zhang, Loakman, Lin, and Guerin]{tang-etal-2023-enhancing}
Chen Tang, Hongbo Zhang, Tyler Loakman, Chenghua Lin, and Frank Guerin.
\newblock Enhancing dialogue generation via dynamic graph knowledge aggregation.
\newblock In \emph{Proceedings of the 61st Annual Meeting of the Association for Computational Linguistics (Volume 1: Long Papers)}, pages 4604--4616, Toronto, Canada, 2023. Association for Computational Linguistics.

\bibitem[Touvron et~al.(2024)]{llama3}
Hugo Touvron et~al.
\newblock The llama 3 herd of models.
\newblock \emph{arXiv preprint arXiv:2407.21783}, 2024.

\bibitem[Wang et~al.(2024{\natexlab{a}})Wang, Ni, Liu, Lu, Chen, Feng, Wei, Qu, Alinejad-Rokny, Lin, and Yang]{Wang2024AutoPatentAM}
Qiyao Wang, Shiwen Ni, Huaren Liu, Shule Lu, Guhong Chen, Xi Feng, Chi Wei, Qiang Qu, Hamid Alinejad-Rokny, Yuan Lin, and Min Yang.
\newblock Autopatent: A multi-agent framework for automatic patent generation.
\newblock \emph{ArXiv}, abs/2412.09796, 2024{\natexlab{a}}.

\bibitem[Wang et~al.(2024{\natexlab{b}})Wang, Guo, Yao, Zhang, Zhang, Wu, Zhang, Dai, Zhang, Wen, Ye, Zhang, and Zhang]{Wang2024AutoSurveyLL}
Yidong Wang, Qi Guo, Wenjin Yao, Hongbo Zhang, Xin Zhang, Zhen Wu, Meishan Zhang, Xinyu Dai, Min Zhang, Qingsong Wen, Wei Ye, Shikun Zhang, and Yue Zhang.
\newblock Autosurvey: Large language models can automatically write surveys.
\newblock \emph{ArXiv}, abs/2406.10252, 2024{\natexlab{b}}.

\bibitem[Wang et~al.(2024{\natexlab{c}})Wang, Wu, Wei, Jegelka, and Wang]{Wang2024ATU}
Yifei Wang, Yuyang Wu, Zeming Wei, Stefanie Jegelka, and Yisen Wang.
\newblock A theoretical understanding of self-correction through in-context alignment.
\newblock \emph{ArXiv}, abs/2405.18634, 2024{\natexlab{c}}.

\bibitem[Wang et~al.(2024{\natexlab{d}})Wang, Zeng, Liu, Meng, Zhou, and Zhang]{Wang2024TasTeTL}
Yutong Wang, Jiali Zeng, Xuebo Liu, Fandong Meng, Jie Zhou, and Min Zhang.
\newblock Taste: Teaching large language models to translate through self-reflection.
\newblock In \emph{Annual Meeting of the Association for Computational Linguistics}, 2024{\natexlab{d}}.

\bibitem[Wen et~al.(2023)Wen, Wang, and Sun]{Wen2023MindMapKG}
Yilin Wen, Zifeng Wang, and Jimeng Sun.
\newblock Mindmap: Knowledge graph prompting sparks graph of thoughts in large language models.
\newblock In \emph{Annual Meeting of the Association for Computational Linguistics}, 2023.

\bibitem[Wu et~al.(2024)Wu, Hee, Hu, and Lee]{wu2024spinning}
Yuhao Wu, Ming~Shan Hee, Zhiqing Hu, and Roy Ka-Wei Lee.
\newblock Spinning the golden thread: Benchmarking long-form generation in language models.
\newblock \emph{arXiv preprint arXiv:2409.02076}, 2024.

\bibitem[Xi et~al.(2025)Xi, Yin, Fang, Wu, Fang, Zhang, Yong, Xie, Huang, and Chen]{Xi2025OmniThinkEK}
Zekun Xi, Wenbiao Yin, Jizhan Fang, Jialong Wu, Runnan Fang, Ningyu Zhang, Jiang Yong, Pengjun Xie, Fei Huang, and Huajun Chen.
\newblock Omnithink: Expanding knowledge boundaries in machine writing through thinking.
\newblock \emph{ArXiv}, abs/2501.09751, 2025.

\bibitem[Yan et~al.(2024)Yan, Zhu, Wang, Gui, and He]{Yan2024MirrorAM}
Hanqi Yan, Qinglin Zhu, Xinyu Wang, Lin Gui, and Yulan He.
\newblock Mirror: A multiple-perspective self-reflection method for knowledge-rich reasoning.
\newblock \emph{ArXiv}, abs/2402.14963, 2024.

\bibitem[Zhang et~al.(2024)Zhang, Madaan, Gao, Zheng, Mishra, Yang, Tandon, and Alon]{Zhang2024InContextPL}
Tianjun Zhang, Aman Madaan, Luyu Gao, Steven Zheng, Swaroop Mishra, Yiming Yang, Niket Tandon, and Uri Alon.
\newblock In-context principle learning from mistakes.
\newblock \emph{ArXiv}, abs/2402.05403, 2024.

\bibitem[Zhou et~al.(2024)Zhou, Liu, Srivastava, Mei, and Tan]{Zhou2024HypothesisGW}
Yangqiaoyu Zhou, Haokun Liu, Tejes Srivastava, Hongyuan Mei, and Chenhao Tan.
\newblock Hypothesis generation with large language models.
\newblock \emph{ArXiv}, abs/2404.04326, 2024.

\end{thebibliography}
}

\end{document}